\documentclass{article}

\usepackage[dblblindworkshop, final]{neurips_2025}
\usepackage[utf8]{inputenc} % allow utf-8 input
\usepackage[T1]{fontenc}    % use 8-bit T1 fonts
\usepackage{hyperref}       % hyperlinks
\usepackage{url}            % simple URL typesetting
\usepackage{booktabs}       % professional-quality tables
\usepackage{amsfonts}       % blackboard math symbols
\usepackage{nicefrac}       % compact symbols for 1/2, etc.
\usepackage{microtype}      % microtypography
\usepackage{xcolor}         % colors

\usepackage{graphicx} 
\usepackage{multirow} 
\usepackage{xcolor}
\usepackage{colortbl}
\usepackage{booktabs,array,ragged2e,tabularx,makecell,xcolor,colortbl}
\newcolumntype{L}[1]{>{\RaggedRight\arraybackslash}m{#1}} % fixed-width, ragged-right
\newcolumntype{Y}{>{\RaggedRight\arraybackslash}X}        % stretchable, ragged-right
\newcommand{\metric}[2]{\makecell[l]{\textbf{#1}\\\textbf{#2}}} % two-line metric label

\usepackage{xcolor}
\usepackage{soul}

\colorlet{hlred}{red!20}
\colorlet{hlyellow}{yellow!30}
\colorlet{hlgreen}{green!20}

\newcommand{\hlc}[2]{\begingroup\sethlcolor{#1}\hl{#2}\endgroup}

\newcommand{\myredhl}[1]{\hlc{hlred}{#1}}
\newcommand{\myyellowhl}[1]{\hlc{hlyellow}{#1}}
\newcommand{\mygreenhl}[1]{\hlc{hlgreen}{#1}}

\title{SmartWilds: Multimodal Wildlife Monitoring Dataset}

\author{%
Jenna Kline \thanks{Corresponding author: kline.377@osu.edu}\And
 Anirudh Potlapally \And
 Bharath Pillai \And
 Tanishka Wani \And
 Vedant Patil \And
 Penelope Covey \And
 Rugved Katole \And
 Samuel Stevens \And
 Hari Subramoni \And
 Tanya Berger-Wolf \And Christopher Stewart \And
 \\
 Department of Computer Science and Engineering \\
 The Ohio State University \\
 Columbus, OH 43201 \\
}

\begin{document}

\maketitle

\begin{abstract}
  We present the first release of SmartWilds, a multimodal wildlife monitoring dataset. 
SmartWilds is a synchronized collection of drone imagery, camera trap photographs and videos, and bioacoustic recordings collected during summer 2025 at The Wilds safari park in Ohio. 
This dataset supports multimodal AI research for comprehensive environmental monitoring, addressing critical needs in endangered species research, conservation ecology, and habitat management.
Our pilot deployment captured four days of synchronized monitoring across three modalities in a 220-acre pasture containing Pere David's deer, Sichuan takin, Przewalski's horses, as well as species native to Ohio. 
We provide a comparative analysis of sensor modality performance, demonstrating complementary strengths for landuse patterns, species detection, behavioral analysis, and habitat monitoring. 
This work establishes reproducible protocols for multimodal wildlife monitoring while contributing open datasets to advance conservation computer vision research. 
Future releases will include synchronized GPS tracking data from tagged individuals, citizen science data, and expanded temporal coverage across multiple seasons.
\end{abstract}

\section{Introduction}

Conservation biology requires comprehensive ecosystem monitoring to inform evidence-based management decisions, yet traditional approaches provide fragmented views of wildlife activity and habitat use.
The integration of multiple sensing modalities, powered by edge AI and computer vision approaches offers transformative opportunities for automated environmental monitoring at unprecedented scales \citep{bessontoward2022, tuiaperspectives2022, pringleopportunities2025, kline2025studying}.
Our research advances multimodal AI for wildlife monitoring by creating datasets and models that enable object detection and tracking across environmental conditions, support fine-grained species classification from multi-sensor data, and facilitate behavioral analysis from long video sequences.

We present a multimodal dataset from sensor deployments at The Wilds Conservation Center \citep{thewilds_website} \footnote{\url{https://imageomics.github.io/naturelab/}}. Our dataset was curated to evaluate multi-sensor fusion techniques and benchmark machine learning approaches for conservation applications across visual, acoustic, and environmental data streams.
This dataset poses challenges to existing computer vision methods through its combination of rare species detection, variable conditions, and the need for behavioral state recognition over extended time periods.
This work supports research areas including endangered species monitoring, conservation and restoration ecology, and habitat assessment for both exotic endangered species and native biodiversity conservation.
Our contributions include: (1) a synchronized multimodal ecological dataset with comprehensive metadata, (2) reproducible protocols for environmental monitoring sensor networks, and (3) support for sensor fusion to advance multimodal learning research in environmental contexts. 

The rest of the paper is organized as follows: Section \ref{relatedwork} reviews related works on multi-modal sensor fusion for ecological research; Section \ref{datacollection} describes the study site and sensor network design; Section \ref{data} details the dataset, including camera trap, bioacoustic, drone, and metadata. Section \ref{discussion} includes field observation, discusses the strengths and weaknesses of the sensing modalities and how the pilot study informed additional data collection; Section \ref{future} includes details about future data releases and work in progress.

\begin{figure}
    \centering
    \includegraphics[width=1\linewidth]{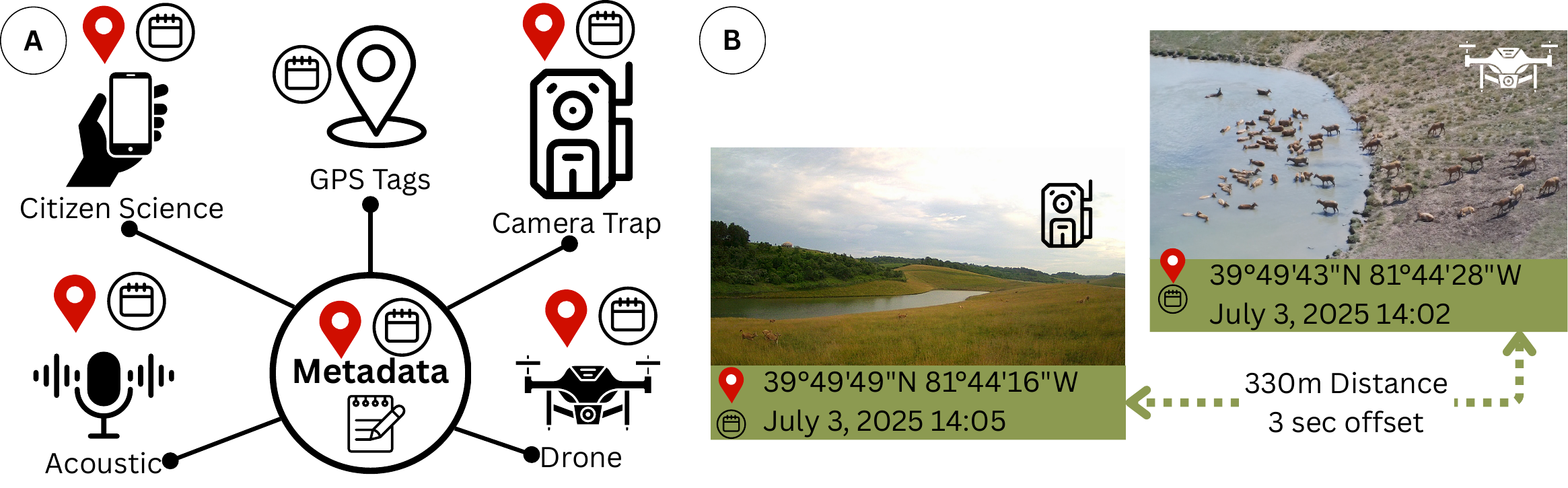}
    \caption{Representative images and data study design. GPS and time-stamp metadata allow for cross-referencing between modalities. (a) Diagram of dataset modalities, citizen science images, GPS tags, acoustic data, camera trap and drone images, joined via location and time-stamp metadata. (b) Example of multi-modal data cross-referencing using metadata. Camera trap view (TW02) of the Pere David's deer synchronized with the drone image of the Pere David's deer wading in the lake.}
    \label{fig:pdeer}
\end{figure}

\section{Related Work}
\label{relatedwork}

Recent advances have introduced multimodal wildlife monitoring datasets addressing diverse conservation challenges \citep{smithobserving2021, buxtonpairiningcamera2018, bessontoward2022}. MammAlps \citep{gabeff2025mammalps} combines multi-view video with synchronized audio data to analyze wild mammal behaviors in alpine environments, demonstrating the power of multimodal approaches for detailed behavioral analysis. BuckTales \citep{naik2024bucktales} provides multi-UAV tracking data for wild antelope identification and re-identification, advancing techniques for individual animal monitoring. The PanAf-FGBG dataset \citep{brookes2025panaf} explores how environmental backgrounds impact wildlife behavior recognition, while YOLO-Behavior \citep{chan2025yolo} demonstrates automated behavior quantification frameworks. KABR \citep{kholiavchenko2024kabr} focuses on Kenyan wildlife behavior recognition from drone footage, contributing to conservation AI in African ecosystems.
However, most existing datasets focus on specific taxonomic groups or behavioral tasks rather than comprehensive ecosystem monitoring. Our work contributes synchronized multi-sensor data collection protocols designed specifically for conservation digital twin development, with planned integration of GPS tracking data for individual-level behavioral analysis.

\section{Field Deployment and Data Collection}
\label{datacollection}
\subsection{Study Site}
\begin{figure}
    \centering
    \includegraphics[width=1\linewidth]{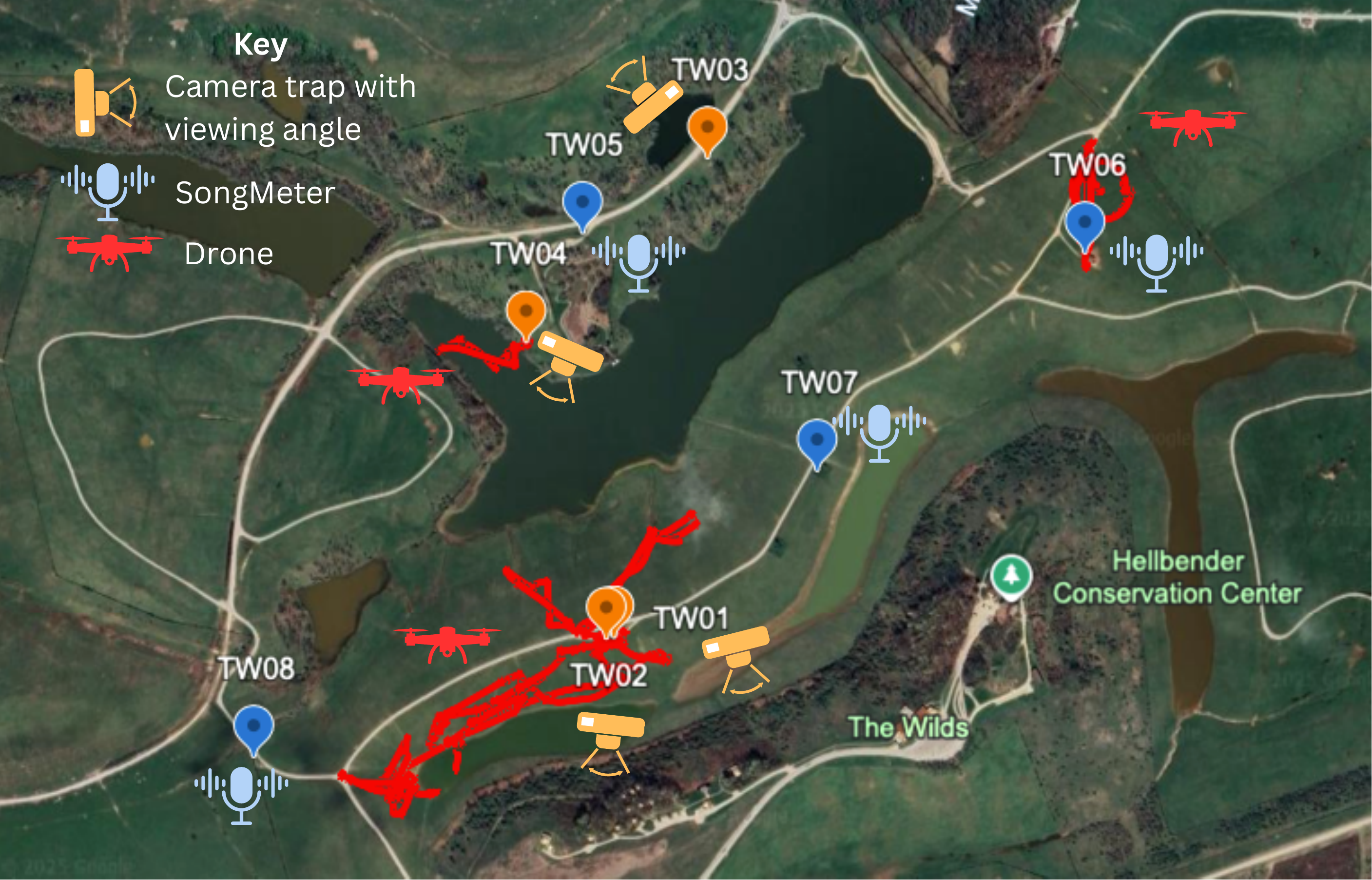}
    \caption{Map of sensor placements created with Google Earth. Camera trap locations in orange, bioacoustics sensors in blue, drone flight paths in red. See Table \ref{tab:site_selection} for site details.}
    \label{fig:map}
\end{figure}

The Wilds is a 10,000-acre conservation center in southeastern Ohio, home to endangered species conservation programs and native wildlife restoration efforts \citep{thewilds_website}. Our pilot deployment focused on a single 220-acre enclosure containing a breeding population of GPS-tagged Pere David's deer (\textit{Elaphurus davidianus}), chosen specifically to enable future integration with individual tracking data. Sensor sites were selected chronologically based on observed wildlife activity patterns and strategic coverage of diverse habitat types within the study pasture, illustrated in Fig. \ref{fig:map} and described in Table \ref{tab:site_selection}. Camera trap sites prioritized high deer activity areas, particularly around water sources, while bioacoustic monitors targeted diverse acoustic environments from open grasslands to woodland edges. Where possible, existing structures were utilized for sensor mounting to minimize environmental impact and maximize equipment protection.

\subsection{Sensor Network Design}
The multimodal sensor network consisted of three complementary sensing technologies deployed for four days of continuous monitoring (June 30 - July 3 2025). Four camera traps, including GardePro T5NG and comparable trail camera models, were strategically positioned around lakes and wildlife congregation areas using motion-triggered photo/video hybrid mode to capture animal activity at key locations. See Table \ref{tab:site_selection} for details on site selection. Four bioacoustic monitors (Song Meter Mini devices) recorded high-quality 48kHz, 16-bit mono audios. 
%
%Half of the SongMeters were configured with scheduled intervals targeting sunrise and sunset periods when bird vocalizations are most active while the other half were configured to record 5 minutes every hour to detect ungulate vocalizations. 
%
Multiple drone missions using Parrot ANAFI quadcopters provided flexible aerial coverage through systematic surveys and opportunistic behavioral tracking, with dedicated synchronization flights conducted within view of camera traps to enable precise cross-modal timestamp calibration. This integrated sensor array enabled comprehensive multimodal data collection across fixed monitoring locations and dynamic aerial observations.

\section{Dataset}
\label{data}
The multi-modal dataset is summarized in Table \ref{data}.
The initial dataset release totals 101GB and over 20K files. The dataset is organized by sensor type and deployment location, with standardized metadata for each component. 
\subsection{Camera trap data}
Camera trap data consists of motion-triggered images and videos organized by deployment site (TW01-TW04), providing comprehensive visual documentation specifically structured for object detection, tracking, and fine-grained classification of wildlife species. The systematic capture of animal behaviors and interactions across multiple camera locations enables localization and recognition tasks in naturalistic settings with varying environmental conditions and species compositions. 
\subsection{Bioacoustic data}
Bioacoustic recordings from continuous and scheduled monitoring (TW05-TW08) were collected to enable multi-sensor fusion research that combines acoustic representations with visual data streams. Half of the monitors were configured to record 5 minutes every hour to capture ungulate vocalizations throughout the day, and half were configured to record bird song at dusk and dawn to capture local diversity.
\subsection{Drone data}
Drone mission data includes video footage with flight telemetry and detailed mission objectives, designed to support 3D modeling, temporal and behavioral reasoning from video sequences. The extended drone video captures are particularly valuable for recognition of complex behavioral states in long video sequences, including territorial displays, social interactions, and habitat use patterns that unfold over extended observation periods.
\subsection{Metadata}
All deployments include comprehensive metadata with GPS coordinates, habitat descriptions, technical sensor specifications, deployment timestamps, environmental conditions, and detailed field observations from researchers. This structured field note documentation, combined with visual and acoustic data streams, creates a rich foundation for multi-sensor fusion studies that integrate images, sounds, and contextual field observations. The standardized metadata framework supports human-in-the-loop and citizen-science annotation efforts by providing the contextual information necessary for active-learning pipelines that balance annotation cost and data quality.

\begin{table}[t]
\centering
\footnotesize
\caption{The Wilds Multimodal Initial Release Dataset Summary. Initial release includes photos, videos, and acoustic data. Future releases will include GPS tag data, citizen science images, and weather and satellite data.}
\label{tab:wilds_dataset}
\begin{tabular}{p{2cm}p{5cm}p{4cm}p{1cm}}
\toprule
\textbf{Modality} & \textbf{Data Type} & \textbf{Total Files} & \textbf{GB} \\
\midrule
Camera Traps & Visual monitoring (photos and videos) & 20,014 & 49 \\
Bioacoustic & Audio recordings & 311 & 6 \\
Drone & Aerial video & 20 video files + metadata & 46 \\
\midrule
\textbf{Total} & \textbf{All modalities} & $\sim$\textbf{20K} & \textbf{101} \\
\bottomrule
\end{tabular}
\end{table}

% Total dataset: 101 GB across all three modalities
% Camera traps: 49 GB despite having 98\% of the files - indicates mostly smaller image files with some videos
% Drone videos: 46 GB for just 20 video files, averaging ~2.3 GB per video)
% Bioacoustics: 6 GB for 311 audio files - indicates substantial audio recordings (averaging ~19 MB per file)

% The storage distribution (49\% camera traps, 46\% drones, 6\% bioacoustics) is much more balanced than the file count distribution, highlighting how the drone videos are particularly data-intensive despite being few in number. This suggests the drone missions captured extensive high-resolution footage, making them valuable for detailed aerial analysis despite the limited number of flights.
% The bioacoustics files, while representing only 6\% of storage, likely contain substantial temporal coverage given their scheduled recording approach, making them efficient for acoustic monitoring relative to storage requirements.

\section{Discussion}
\label{discussion}

\subsection{Field Observations}
Field deployment revealed important insights about multimodal monitoring in conservation settings. Animal responses varied by sensor type. The deer initially showed curiosity toward drone flights but exhibited minimal behavioral disruption overall. Breeding season activity patterns were clearly observable, with territorial males vocalizing frequently and herds congregating around water sources during warm weather. Technical challenges included GPS signal limitations in remote areas affecting some sensor synchronization, weather impacts on acoustic recording quality, and the need for creative mounting solutions in areas lacking suitable structures. Despite these challenges, the sensor network successfully captured multimodal data across all target areas.

\subsection{Comparison of Sensor Modalities}
We summarize the relative performance each sensing modality across eight key performance dimensions relevant to conservation monitoring applications (Table \ref{tab:sensor_comparison}). These metrics were selected based on established frameworks for conservation technology evaluation and practical deployment considerations in wildlife monitoring contexts.
Spatial range and resolution determines monitoring coverage and detection capabilities across different habitat scales \citep{tuiaperspectives2022, pringleopportunities2025}. Temporal range and resolution captures both short-term behavioral events and long-term ecological patterns \cite{bessontoward2022}. Species detectability captures different sensor modalities ability to sense specific taxonomic groups, especially more cryptic species such as birds or insects,\citep{smithobserving2021}. Behavioral detail is important for understanding complex social interactions of group-living animals and individual to group-level responses to environmental changes \citep{kline2025studying}. Deployment effort and data volume captures practical considerations affecting scalability and cost of long-term monitoring efforts \citep{bessontoward2022}.

\begin{table*}[t]
\centering
\caption{Comparative analysis of sensor modality performance across key conservation monitoring metrics. 
Metrics selected based on established frameworks for wildlife monitoring technology evaluation \citep{tuiaperspectives2022, bessontoward2022} and practical deployment considerations in conservation settings. Performance rating: \myredhl{Poor}, \myyellowhl{Moderate}, \mygreenhl{Good}. \\
\footnotesize *GPS tag data will be added in a future data release.}
\label{tab:sensor_comparison}

% tighter but readable spacing
\setlength{\tabcolsep}{4.5pt}
\renewcommand{\arraystretch}{1.15}

\footnotesize
\begin{tabularx}{\textwidth}{L{1.8cm} Y Y Y Y}
\toprule
\textbf{Metric} & \textbf{Camera Traps} & \textbf{Bioacoustics} & \textbf{Drones} & \textbf{GPS Tags*} \\
\midrule
\metric{Spatial}{Range}
  & \cellcolor{red!20}Fixed location, $\sim$30\,m radius
  & \cellcolor{red!20}Fixed location, $\sim$100\,m radius
  & \cellcolor{yellow!30}Mobile; battery-limited ($\sim$2\,km)
  & \cellcolor{green!20}Entire home range \\
\metric{Spatial}{Resolution}
  & \cellcolor{green!20}High within field-of-view
  & \cellcolor{yellow!30}Moderate directional
  & \cellcolor{green!20}Sub-meter aerial resolution
  & \cellcolor{yellow!30}$\sim$1--10\,m accuracy \\
\metric{Temporal}{Range}
  & \cellcolor{green!20}Weeks to months
  & \cellcolor{green!20}Weeks to months
  & \cellcolor{red!20}Hours per mission
  & \cellcolor{green!20}Months to years \\
\metric{Temporal}{Resolution}
  & \cellcolor{green!20}Event-triggered; $<\!1$\,s
  & \cellcolor{green!20}Continuous or scheduled
  & \cellcolor{green!20}30--60\,fps video
  & \cellcolor{red!20}Hourly locations \\
\metric{Species}{Detectability}
  & \cellcolor{yellow!30}Large ungulates, visible species
  & \cellcolor{green!20}Cryptic/vocal species, birds
  & \cellcolor{yellow!30}Large mammals, aerial view
  & \cellcolor{red!20}Tagged individuals only \\
\metric{Behavior}{Detail}
  & \cellcolor{red!20}Limited to frame interactions
  & \cellcolor{yellow!30}Vocalizations, acoustic behaviors
  & \cellcolor{green!20}High detail: posture, interactions
  & \cellcolor{red!20}Movement patterns only \\
\metric{Deployment}{Effort}
  & \cellcolor{green!20}Low--medium (site visits)
  & \cellcolor{green!20}Low--medium (site visits)
  & \cellcolor{red!20}High (active piloting)
  & \cellcolor{green!20}Low once deployed \\
\metric{Data}{Volume}
  & \cellcolor{green!20}Moderate
  & \cellcolor{yellow!30}Moderate--high
  & \cellcolor{red!20}High
  & \cellcolor{green!20}Low \\
\bottomrule
\end{tabularx}
\end{table*}

\section{Future Directions}
\label{future}

Building on pilot deployment insights, upcoming releases will address identified limitations and leverage demonstrated multimodal strengths. The pilot revealed minimal useful data from camera trap videos compared to drone footage, leading to modified protocols with co-located bioacoustic monitors and camera traps at three additional sites for direct detection capability comparison. Future releases will integrate synchronized GPS tracking from ear-tagged Pere David's deer with visual and acoustic data, enabling analysis of individual movement patterns and behaviors across the four additional weeks of planned data collection to capture seasonal variation.

The demonstrated complementary strengths across modalities—where camera traps excel at species identification, bioacoustic monitors provide continuous temporal coverage, and drones offer landscape-scale perspectives. These complementary strengths will inform machine learning research directions focused on multimodal fusion architectures. Development will prioritize real-time adaptive sampling through edge computing capabilities and AI-assisted management systems that leverage integrated sensor networks' superior performance over single-modality approaches. Extension to multiple habitat types, replication at additional conservation sites, and integration of citizen science observations will expand data collection while validating multimodal AI frameworks that can operate autonomously across diverse ecosystems and transform global environmental monitoring practices.

\section{Data Availability Statement}
 The dataset is freely available on \href{https://huggingface.co/collections/imageomics/smartwilds-6880ec9381da004aa5682cf6}{HuggingFace} under a CC0-1.0 license. The dataset is arranged by modality, including \href{https://huggingface.co/datasets/imageomics/thewilds_cameratraps}{camera trap images}, \href{https://huggingface.co/datasets/imageomics/thewilds_bioacousticmonitors}{bioacoustic recordings}, and \href{https://huggingface.co/datasets/imageomics/thewilds_drones}{drone video footage}.

Future releases will include GPS tracking data of Pere David's deer, citizen science images captured using mobile phones and cameras, and detailed weather and habitat data from satellite data sources. We also plan to expand citizen science integration through crowdsourced imagery validation and expert-guided annotation workflows.

\begin{ack}
We thank Dan Beetem, The Wilds, and the Columbus Zoo \& Aquarium for their support in facilitating this project. All data collection, including drone flights, was conducted under the supervision of the Director of Animal Management, with permission from The Wilds Animal Care and Use Committee. \\
\\
This project is supported by the \href{https://icicle.osu.edu/}{AI Institute for Intelligent Cyberinfrastructure with Computational Learning in the Environment (ICICLE)}, the \href{https://imageomics.org/}{Imageomics Institute}, and the \href{http://abcresearchcenter.org/}{AI and Biodiversity Change (ABC) Global Center}. ICICLE is funded by the US National Science Foundation under \href{https://www.nsf.gov/awardsearch/showAward?AWD_ID=2112606}{Award No. 2112606}, the Imageomics Institute is funded by the US National Science Foundation's Harnessing the Data Revolution (HDR) program under \href{https://www.nsf.gov/awardsearch/showAward?AWD_ID=2118240}{Award No. 2118240} (Imageomics: A New Frontier of Biological Information Powered by Knowledge-Guided Machine Learning). The ABC Global Center is funded by the US National Science Foundation under \href{https://www.nsf.gov/awardsearch/showAward?AWD_ID=2330423&HistoricalAwards=false}{Award No. 2330423} and the Natural Sciences and Engineering Research Council of Canada under \href{https://www.nserc-crsng.gc.ca/ase-oro/Details-Detailles_eng.asp?id=782440}{Award No. 585136}. This work draws on research supported by the Social Sciences and Humanities Research Council.
\end{ack}

\bibliographystyle{apalike}
\bibliography{ref}

@article{gabeff2025mammalps,
  title={MammAlps: A multi-view video behavior monitoring dataset of wild mammals in the Swiss Alps},
  author={Gabeff, V. and Qi, H. and Flaherty, B. and others},
  year={2025}
}

@article{naik2024bucktales,
  title={BuckTales: A multi-UAV dataset for multi-object tracking and re-identification of wild antelopes},
  author={Naik, Hemal and Yang, Junran and Das, Dipin and Crofoot, Margaret and Rathore, Akanksha and Sridhar, Vivek Hari},
  journal={Advances in Neural Information Processing Systems},
  volume={37},
  pages={81992--82009},
  year={2024}
}

@article{brookes2025panaf,
  title={The PanAf-FGBG Dataset: Understanding the Impact of Backgrounds in Wildlife Behaviour Recognition},
  author={Brookes, O. and Kukushkin, M. and Mirmehdi, M. and others},
  year={2025}
}

@article{chan2025yolo,
  title={YOLO-Behaviour: A simple, flexible framework to automatically quantify animal behaviours from videos},
  author={Chan, A. H. H. and Putra, P. and Schupp, H. and others},
  journal={Methods in Ecology and Evolution},
  volume={16},
  pages={760--774},
  year={2025}
}

@article{kholiavchenko2024kabr,
  title={KABR: In-Situ Dataset for Kenyan Animal Behavior Recognition From Drone Videos},
  author={Kholiavchenko, M. and Kline, J. and Ramirez, M. and others},
  year={2024}
}

@article{bessontoward2022, title={Towards the fully automated monitoring of ecological communities}, volume={25}, ISSN={1461-0248}, DOI={10.1111/ele.14123}, abstractNote={High-resolution monitoring is fundamental to understand ecosystems dynamics in an era of global change and biodiversity declines. While real-time and automated monitoring of abiotic components has been possible for some time, monitoring biotic components—for example, individual behaviours and traits, and species abundance and distribution—is far more challenging. Recent technological advancements offer potential solutions to achieve this through: (i) increasingly affordable high-throughput recording hardware, which can collect rich multidimensional data, and (ii) increasingly accessible artificial intelligence approaches, which can extract ecological knowledge from large datasets. However, automating the monitoring of facets of ecological communities via such technologies has primarily been achieved at low spatiotemporal resolutions within limited steps of the monitoring workflow. Here, we review existing technologies for data recording and processing that enable automated monitoring of ecological communities. We then present novel frameworks that combine such technologies, forming fully automated pipelines to detect, track, classify and count multiple species, and record behavioural and morphological traits, at resolutions which have previously been impossible to achieve. Based on these rapidly developing technologies, we illustrate a solution to one of the greatest challenges in ecology: the ability to rapidly generate high-resolution, multidimensional and standardised data across complex ecologies.}, number={12}, journal={Ecology Letters}, author={Besson, Marc and Alison, Jamie and Bjerge, Kim and Gorochowski, Thomas E. and Høye, Toke T. and Jucker, Tommaso and Mann, Hjalte M. R. and Clements, Christopher F.}, year={2022}, pages={2753–2775}, language={en} }

@article{buxtonpairiningcamera2018, title={Pairing camera traps and acoustic recorders to monitor the ecological impact of human disturbance}, volume={16}, rights={https://www.elsevier.com/tdm/userlicense/1.0/}, ISSN={23519894}, DOI={10.1016/j.gecco.2018.e00493}, abstractNote={Over the past two decades, the use of camera traps and acoustic monitoring in the investigation of animal ecology have grown rapidly, with each technique enhancing broadscale wildlife surveying. Camera traps are a cost-effective, noninvasive means of sampling communities of mid-to large-terrestrial species, and acoustic recording devices capture human sounds and sound-producing animals, including species of mammals, birds, anurans, and insects. Rarely are these techniques combined, despite the advantages of merging their respective strengths. Namely, camera traps paired with acoustic recorders can evaluate the abundance, distribution, and behavior of multiple guilds and trophic levels across landscapes while concurrently monitoring multiple human stressors in real time. Moreover, integrating these approaches enhances detection accuracy and strengthens statistical inference at multiple survey scales. We conducted a literature review, and found only 13 studies that combine camera traps and acoustic recorders, 8 of which either compared the ability of each technique to detect species of interest or discussed the advantages of each technique. We outline potential questions that can be addressed by pairing acoustic recorders and camera traps, including enabling the simultaneous assessment of noise pollution and its impacts on mammal and avian communities. Furthermore, we discuss how the analysis of data from each technique face similar challenges; thus, simultaneous innovation offers the ability to apply solutions to both techniques and amplify their respective strengths. Digital technologies and big data are changing nature conservation in increasingly profound ways and integration of camera traps and acoustic recorders will facilitate new, transformative discoveries to meet modern conservation challenges.}, journal={Global Ecology and Conservation}, author={Buxton, Rachel T. and Lendrum, Patrick E. and Crooks, Kevin R. and Wittemyer, George}, year={2018}, month=oct, pages={e00493}, language={en} }

@article{pringleopportunities2025, title={Opportunities and challenges for monitoring terrestrial biodiversity in the robotics age}, ISSN={2397-334X}, url={https://kar.kent.ac.uk/109784/}, abstractNote={With biodiversity loss escalating globally, a step-change is needed in our capacity to accurately monitor species populations across ecosystems. Robotic and autonomous systems (RAS) offer technological solutions that may significantly advance terrestrial biodiversity monitoring, but this potential is yet to be considered systematically. We used a modified Delphi technique to synthesise knowledge from 98 biodiversity and 31 RAS experts who identified the major methodological barriers that currently hinder monitoring, and explored the opportunities and challenges that RAS offer to overcome these barriers. Biodiversity experts identified four barrier categories: site access, species/individual identification, data handling/storage and power/network availability. Robotics experts highlighted technologies that could overcome these barriers and identified the developments needed to facilitate RAS-based autonomous biodiversity monitoring. Some existing RAS could be optimised relatively easily to survey species, but would require development to monitor more ‘difficult’ taxa and be robust enough to work in uncontrolled conditions within ecosystems. Other nascent technologies (e.g., novel sensors, biodegradable robots) need accelerated research. Overall, it was felt that RAS could lead to major progress in monitoring terrestrial biodiversity by supplementing, rather than supplanting, existing methods. Transdisciplinarity needs to be fostered between biodiversity and RAS experts, so future ideas and technologies can be co-developed effectively.}, note={Accepted: 2025-04-07}, journal={Nature Ecology and Evolution}, publisher={Springer Nature}, author={Pringle, Stephen and Dallimer, Martin and Goddard, Mark A. and Le Goff, Léni E. and Hart, Emma and Langdale, Simon J. and Fisher, Jessica C. and Abad, Sara-Adela and Ancrenaz, Marc and Angeoletto, Fabio and Auat Cheein, Fernando and Austen, Gail E. and Bailey, Joseph and Baldock, Katherine and Banin, Lindsay and Banks-Leite, Cristina and Barau, Aliyu and Bashyal, Reshu and Bates, Adam J. and Bicknell, Jake E. and Bielby, Jon and Bosilj, Petra and Bush, Emma and Butler, Simon and Carpenter, Dan and Clements, Christopher F. and Cully, Antoine and Davies, Kendi and Deere, Nicolas J. and Dodd, Michael and Drinkwater, Rosie and Driscoll, Don and Dutilleux, Guillaume and Dyrmann, Mads and Edwards, David P. and Farhadinia, Mohammad S. and Faruk, Aisyah and Field, Richard and Fletcher, Robert J. and Foster, Chris and Fox, Richard and Francksen, Richard and Franco, Aldina and Gainsbury, Alison and Gardner, Charlie J. and Giogi, Ioanna and Griffiths, Richard A. and Hamaza, Salua and Hanheide, Marc and Hayward, Matt W. and Hedblom, Marcus and Helgason, Thorunn and Heon, Sui P. and Hughes, Kevin and Hunt, Edmund and Ingram, Daniel J. and Jackson-Mills, George and Jowett, Kelly and Keitt, Timothy and Kloepper, Laura and Kramer-Schadt, Stephanie and Labisko, Jim and Labrosse, Frédéric and Lawson, Jenna and Lecomte, Nicolas and de Lima, Ricardo F. and Littlewood, Nick A. and Marshall, Harry and Masala, Giovanni Luca and Maskell, Lindsay and Matechou, Eleni and Mazzolai, Barbara and McConnell, Alistair and Melbourne, Brett and Miriyev, Aslan and Nana, Eric and Ossola, Alessandro and Papworth, Sarah and Parr, Catherine and Payo-Payo, Ana and Perry, Gad and Pettorelli, Nathalie and Pillay, Rajeev and Potts, Simon G. and Prendergast-Miller, Miranda and Qie, Lan and Rolley-Parnell, Persie and Rossiter, Stephen J. and Rowcliffe, J. Marcus and Rumble, Heather and Sadler, Jon P. and Sandom, Christopher and Sanyal, Asiem and Schrodt, Franziska and Sethi, Sarab S. and Shabrani, Adi and Siddall, Robert and Smith, Simón and Snep, Robbert P. H. and Soulsbury, Carl D. and Stanley, Margaret C. and Stephens, Philip A. and Stephenson, P. J. and Struebig, Matthew J. and Studley, Matthew and Svátek, Martin and Tang, Gilbert and Taylor, Nicholas and Umbers, Kate and Ward, Robert and White, Patrick and Whittingham, Mark J. and Wich, Serge and Williams, Christopher D. and Yoh, Natalie and Zaidi, Syed Ali Raza and Zmarz, Anna and Zwerts, Joeri and Davies, Zoe G.}, year={2025}, month=apr, language={en} }

@article{smithobserving2021, title={Observing the unwatchable: Integrating automated sensing, naturalistic observations and animal social network analysis in the age of big data}, volume={90}, rights={© 2020 British Ecological Society}, ISSN={1365-2656}, DOI={10.1111/1365-2656.13362}, abstractNote={In the 4.5 decades since Altmann (1974) published her seminal paper on the methods for the observational study of behaviour, automated detection and analysis of social interaction networks have fundamentally transformed the ways that ecologists study social behaviour. Methodological developments for collecting data remotely on social behaviour involve indirect inference of associations, direct recordings of interactions and machine vision. These recent technological advances are improving the scale and resolution with which we can dissect interactions among animals. They are also revealing new intricacies of animal social interactions at spatial and temporal resolutions as well as in ecological contexts that have been hidden from humans, making the unwatchable seeable. We first outline how these technological applications are permitting researchers to collect exquisitely detailed information with little observer bias. We further recognize new emerging challenges from these new reality-mining approaches. While technological advances in automating data collection and its analysis are moving at an unprecedented rate, we urge ecologists to thoughtfully combine these new tools with classic behavioural and ecological monitoring methods to place our understanding of animal social networks within fundamental biological contexts.}, number={1}, journal={Journal of Animal Ecology}, author={Smith, Jennifer E. and Pinter-Wollman, Noa}, year={2021}, pages={62–75}, language={en} }

@article{tuiaperspectives2022, title={Perspectives in machine learning for wildlife conservation}, volume={13}, ISSN={2041-1723}, DOI={10.1038/s41467-022-27980-y}, number={1}, journal={Nature Communications}, author={Tuia, Devis and Kellenberger, Benjamin and Beery, Sara and Costelloe, Blair R. and Zuffi, Silvia and Risse, Benjamin and Mathis, Alexander and Mathis, Mackenzie W. and Van Langevelde, Frank and Burghardt, Tilo and Kays, Roland and Klinck, Holger and Wikelski, Martin and Couzin, Iain D. and Van Horn, Grant and Crofoot, Margaret C. and Stewart, Charles V. and Berger-Wolf, Tanya}, year={2022}, month=feb, pages={792}, language={en} }

@article{kline2025studying,
  title={Studying collective animal behaviour with drones and computer vision},
  author={Kline, Jenna and Afridi, Saadia and Rolland, Edouard GA and Maalouf, Guy and Laporte-Devylder, Lucie and Stewart, Christopher and Crofoot, Margaret and Stewart, Charles V and Rubenstein, Daniel I and Berger-Wolf, Tanya},
  journal={Methods in Ecology and Evolution},
  year={2025},
  publisher={Wiley Online Library}
}

@misc{thewilds_website,
  author       = {{The Wilds}},
  title        = {The Wilds: Conservation Center},
  howpublished = {\url{https://www.thewilds.org/}},
  year         = {2025},
  note         = {Accessed: 2025-09-09}
}

%%%%%%%%%%%%%%%%%%%%%%%%%%%%%%%%%%%%%%%%%%%%%%%%%%%%%%%%%%%%

\appendix

\section{Supplemental Material}

\begin{table*}[htbp]
\caption{Sensor deployment sites with selection rationale and habitat characteristics. \\ \footnotesize CT = camera trap. SM = SongMeter bioacoustic recording device).}
\centering
\footnotesize
\begin{tabular}{p{1cm}p{1cm}p{2.5cm}p{2.5cm}p{5cm}}
\toprule
\textbf{Site ID} & \textbf{Sensor Type} & \textbf{Site Name} & \textbf{Habitat Type} & \textbf{Selection Rationale} \\
\midrule 
TW01 & CT & Nomad Ridge Shelter (East) & Elevated structure above lake & High Pere David's deer activity at lake; protected mounting location \\
TW02 & CT & Nomad Ridge Shelter (West) & Elevated structure above lake & Complementary lake coverage; high deer activity observed \\
TW03 & CT & Old Giraffe Feeder (Pasture D) & Feeding structure near lake with salt lick & High deer activity around salt lick; artificial congregation point \\
TW04 & CT & Lake Trail (Northwest facing) & Tree-mounted overlooking lake & High deer activity along lake trail; natural travel corridor \\
TW05 & SM & Lake Trail Gate O & Gate structure with vegetation & Easy maintenance access; high bird activity; sunrise/sunset recording \\
TW06 & SM & Zebra Shelter & Open plains structure & Ungulate activity around shelter; daytime recording schedule \\
TW07 & SM & Zipline Tower & Open field structure & Maintenance accessibility; diverse bird activity; sunrise/sunset recording \\
TW08 & SM & Fence with Dense Vegetation & Pasture edge near lake & Capture different acoustic environment; hourly recording \\
\bottomrule
\end{tabular}
\label{tab:site_selection}
\end{table*}

%%%%%%%%%%%%%%%%%%%%%%%%%%%%%%%%%%%%%%%%%%%%%%%%%%%%%%%%%%%%

\end{document}